\documentclass[11pt,a4paper]{article}

\usepackage[margin=1in]{geometry}
\usepackage{amsmath, amssymb}
\usepackage{graphicx}
\usepackage{booktabs}
\usepackage{hyperref}
\usepackage{cite}
\usepackage{algorithm}
\usepackage{algpseudocode}
\usepackage{xcolor}
\usepackage{titlesec}
\usepackage{abstract}
\usepackage{parskip}
\usepackage{microtype}

\hypersetup{
    colorlinks=true,
    linkcolor=blue,
    citecolor=blue,
    urlcolor=blue
}

\title{
    \noindent\rule{\linewidth}{1.5pt} \\[0.4em]
    \Large \textbf{Not All Transitions Matter: Evidence from PPO} \\[0.3em]
    \large {Reducing Temporal Correlation in PPO Without Degrading Performance} \\[0.4em]
    \noindent\rule{\linewidth}{1.5pt}
}

\author{
\begin{tabular}{c}
\textbf{Ajhesh Basnet} \\
\texttt{ajheshb@gmail.com}\\
Department of Artificial Intelligence and Data Science \\
KPR Institute of Engineering and Technology, Coimbatore
\end{tabular}
}
\date{}

\begin{document}

\maketitle

\begin{abstract}
Training a reinforcement learning agent on-policy means collecting fresh experience at every update, and that experience comes with a hidden problem. Each state in a rollout is the direct output of the previous one, causally chained together by the agent's own actions. Because of this, consecutive transitions are never truly independent. They carry overlapping information, and the gradient signal the network receives ends up far more repetitive than the batch size suggests. The same directions get reinforced over and over, the value network struggles to keep up as the policy shifts, and training becomes quietly unstable in ways that reward curves alone rarely reveal.

This paper asks whether that redundancy can simply be removed. We show that randomly dropping a fixed fraction of transitions from the rollout, at the right stage so the reward signal stays intact, is enough to break the repetitive gradient structure and stabilise training. The change is minimal: one sampling step, no new components, no modification to the core algorithm, and it works with any PPO implementation. Across five environments of increasing difficulty, CartPole-v1, Acrobot-v1, LunarLander-v2, HalfCheetah-v5, and Hopper-v5, the method matches vanilla PPO on reward while producing more consistent training dynamics across KL divergence, policy entropy, and value estimates. Dropping $25\%$ of transitions turns out to be the sweet spot: enough to disrupt the redundancy, not enough to thin the batch. Source code: \url{https://github.com/ajheshbasnet/rollout-slim}.
\end{abstract}

\section{Introduction}

Supervised learning assumes training samples are Independently and Identically Distributed (IID). Under this assumption, gradient signals across batches stay diverse and non-redundant, and learning proceeds stably. In on-policy RL, this assumption does not just get bent. It gets broken by construction.

Within a single trajectory $\tau$, every state $s_{t+1}$ is a direct causal product of $s_t$. The trajectory is not a collection of independent observations but a chain, where each transition follows from the policy's own prior decisions. Feeding this into a neural network produces gradient vectors that point in nearly the same direction update after update, creating functionally collinear weight changes that slow and destabilize learning in ways that are hard to detect until training already looks wrong.

There is a second problem that is harder to see. Once the policy updates past $\tau_1$, the value network starts operating on states it was never trained on. Stale value estimates feed into corrupted advantage signals, which feed into corrupted policy gradients, which push the agent into new regions of state space the critic has seen even less of, and the cycle keeps tightening. This is a direct instance of the Deadly Triad \cite{sutton2018reinforcement}: function approximation combined with bootstrapping under a non-stationary data distribution.

Off-policy methods like DQN and SAC sidestep this through experience replay. Transitions get stored, shuffled, and randomly sampled, which approximates IID well enough in practice. On-policy methods like PPO \cite{schulman2017proximal} cannot do this. The data must come from the current policy, so old transitions cannot be reused, and the temporal correlation problem is left sitting there unaddressed in every rollout.

What makes this worth investigating further is a counterintuitive observation: if you randomly drop a large fraction of transitions from the rollout and still match the performance of the agent that trained on all of them, it strongly implies that most of those transitions were carrying nearly the same gradient signal anyway. The states are so correlated with each other that the network was not really learning from each one independently — it was seeing the same information repeated. That is the core finding this paper builds on.

\paragraph{What about vectorized environments?}
A common practical response to temporal correlation is to run 6--8 parallel environment copies simultaneously, collecting transitions from multiple independent trajectories at once. This does reduce correlation across the batch because each worker is at a different point in a different episode. It works. But running $N$ environments in parallel means $N$ times the memory and $N$ times the CPU overhead, and on a single machine or constrained hardware that cost adds up fast. The method presented here achieves similar decorrelation benefits from a single environment rollout by subsampling after advantage estimation. It is not a replacement for vectorized environments in every setting, but it is a significantly cheaper path to the same goal.

This paper looks at three methods for reducing that correlation without touching the core PPO objective or hurting performance.

\section{Background}

\subsection{On-Policy Trajectory Collection}

In on-policy RL, the agent collects a trajectory by running its current policy $\pi_\theta$ in the environment:
\begin{equation}
\tau = (s_0, a_0, r_0) \rightarrow (s_1, a_1, r_1) \rightarrow (s_2, a_2, r_2) \rightarrow \cdots \rightarrow (s_T, a_T, r_T)
\end{equation}
Every transition $(s_t, a_t, r_t, s_{t+1})$ is causally produced by the one before it, making the trajectory a temporally dependent chain rather than a collection of independent observations.

\subsection{Temporal Correlation and Its Effect on Gradients}

The policy gradient objective is:
\begin{equation}
\nabla J(\theta) = \mathbb{E}_{\tau \sim \pi_\theta} \left[ \sum_t \nabla \log \pi_\theta(a_t | s_t) \cdot A_t \right]
\end{equation}
This expectation assumes samples come from a stationary distribution. In practice, consecutive states $s_t$ and $s_{t+1}$ are highly similar, so gradient vectors $\nabla \log \pi_\theta(a_t|s_t)$ and $\nabla \log \pi_\theta(a_{t+1}|s_{t+1})$ end up nearly parallel.

\subsection{The Non-Stationary Bootstrapping Feedback Loop}

The TD update for the value network is:
\begin{equation}
V(s_t) \leftarrow V(s_t) + \alpha \left[ r_t + \gamma V(s_{t+1}) - V(s_t) \right]
\end{equation}
Both $V(s_t)$ and $V(s_{t+1})$ come from the same network whose weights are shifting during training. After training on $\tau_1$, the critic fits those states well and the loss is low. After $n$ more trajectory updates the weights have drifted substantially. Evaluating the critic on $\tau_1$ now gives high error, even though that error was low at the time of original training. This compounds across trajectories and gives rise to the Non-Stationary Bootstrapping Feedback Loop, a specific case of the Deadly Triad \cite{sutton2018reinforcement}: function approximation plus bootstrapping plus a shifting data distribution.

To see why this is a problem, consider a simple analogy. Imagine a teacher who trains a student exclusively on calculus. Every example, every drill, every test comes from that one chapter. The student performs well on calculus. Then weeks later the teacher starts asking questions from algebra, assuming the student is equally well-prepared. The student struggles, not because they forgot calculus, but because they were never trained on the new material. The value network faces exactly this situation. It is trained on the state distribution of $\tau_1$, but as the policy updates the agent visits different states, and the distribution the critic is asked to evaluate shifts underneath it. The critic's predictions become unreliable not because the network forgot what it learned, but because what it learned no longer matches what it is being asked about.

\subsection{Why On-Policy Methods Must Discard Old Data}

Reusing trajectories from a past policy $\pi_{\theta_{\text{old}}}$ to update the current policy $\pi_\theta$ introduces distribution mismatch. The training data no longer reflects where the current policy actually goes, which biases the gradient in ways that compound over updates. Old trajectories are discarded not as a practical inconvenience but because using them violates the assumptions the objective was derived under.

\subsection{Related Work}

The data efficiency and correlation problems in on-policy RL have attracted a growing body of work, each approaching the issue from a different angle. GePPO \cite{queeney2021geppo} extends PPO to an off-policy setting by deriving policy improvement guarantees that hold under sample reuse, connecting those bounds directly to the clipping mechanism in the original algorithm. PTR-PPO \cite{liang2021ptrppo} takes a complementary approach, combining on-policy collection with prioritized replay of older trajectories to squeeze more signal out of each rollout. PPG \cite{cobbe2021ppg} separates the policy and value optimization phases entirely, allowing the critic to train with higher sample reuse without interfering with the policy's stability. More recently, PROPS \cite{corrado2023props} observed that finite on-policy samples often fail to match the true on-policy distribution — that sampling error itself is a source of high-variance gradients — and addressed this by using an adaptive off-policy behavior policy to collect data that better approximates the current policy's distribution. SAPG \cite{sapg2024} takes a different path: it divides parallel environments into blocks, each optimizing a separate policy, then combines them via an off-policy update to recover data diversity that single-policy rollouts inherently lack.

A separate line of work has studied how the statistical structure of data within a rollout — not just how it is collected — shapes learning. Hollenstein et al.\ \cite{hollenstein2024colorednoise} found that correlated action noise in PPO systematically improves exploration, with the optimal noise color sitting between white and pink noise depending on the amount of data collected per update. Tavakoli et al.\ \cite{tavakoli2021actionredundancy} showed more broadly that action redundancy — where different actions induce nearly identical next-state transitions — is a fundamental problem in RL and degrades sample efficiency in both discrete and continuous settings.

What these methods share is that they all intervene at the data collection stage: they change how trajectories are gathered, how old ones are weighted, or how behavior policies are chosen. Our approach is different. We leave the rollout and advantage estimation completely unchanged and only subsample transitions after the advantage computation is done, at the point where the gradient update is formed. This means the reward signal is fully preserved while the optimization batch becomes less redundant — a distinction that turns out to matter, as Methods 1 and 2 in this paper demonstrate when they fail precisely because they intervene earlier and damage the credit assignment signal in the process.

\section{Methods}

All three methods were evaluated using PPO on five environments: CartPole-v1, Acrobot-v1, LunarLander-v2, HalfCheetah-v5, and Hopper-v5.

\subsection{Method 1: Fixed K-Step Sampling}

Here, transitions are stored in the buffer only once every $K$ steps. Intermediate rewards are not thrown away — they are accumulated and added to the stored transition's reward. For example, with states $s_0, s_1, s_2, s_3, s_4$: $s_0$ is stored, $s_1$ and $s_2$ are skipped but their rewards are accumulated, and $s_3$ is stored with reward $r_1 + r_2 + r_3$. This keeps the total reward signal while reducing the number of correlated transitions in the buffer.

\paragraph{Motivation.} Putting a fixed temporal gap between stored samples and accumulating intermediate rewards is supposed to reduce gradient collinearity without throwing away the reward signal from skipped steps entirely.

\paragraph{Result.} This method works reasonably well only in low-complexity discrete environments like CartPole-v1, where the state space is small, the reward signal is dense and simple, and the fixed skip interval is sufficient to capture the key state transitions. However, it fails in environments of greater complexity. On Acrobot-v1, which has a sparser reward structure and requires coordinated multi-joint control, the method shows early instability. LunarLander-v2 exposes the method's core weakness. Summing rewards over skipped steps loses the fine-grained causal signal — the agent cannot tell what it did right or wrong at any specific timestep, and convergence suffers for it. The fixed skip interval adds another problem on top: it punches the same holes in every trajectory, and those blind spots never rotate out.

\subsection{Method 2: Random Adaptive K-Step Sampling}

Method 1 always skips the same positions — so the fix is straightforward: stop using a fixed interval and draw the skip randomly instead:
\begin{align}
\varepsilon &= \mathcal{N}(0, 1) \\
k' &= \begin{cases} k & \text{if } \varepsilon > 0 \\ k + 1 & \text{if } \varepsilon \leq 0 \end{cases}
\end{align}

\paragraph{Motivation.} A fixed $K$ means the same states get skipped on every trajectory without exception — if $K = 2$, only even-indexed states ever enter the buffer and every odd-indexed state is permanently invisible to the optimizer. Randomizing the skip interval via $\varepsilon \sim \mathcal{N}(0, 1)$ fixes that: the gap shifts each time, so the buffer stops having a fixed parity bias and actually sees a broader, more representative slice of the trajectory over time.

\paragraph{Result.} Method 2 is a genuine improvement over Method 1 — CartPole-v1 trains more cleanly, and on Acrobot-v1 the rotating blind spots help noticeably. But LunarLander-v2 still does not work, and the reason is that randomizing the skip did not touch the actual problem: rewards are still being summed across skipped steps, and that summation is what kills precise credit assignment in environments with shaped rewards. Both methods ultimately share the same flaw — they are fine for small, simple, discrete games, but the moment the environment requires the agent to understand exactly which action caused which outcome, they fall apart.

\subsection{Method 3: Random P\% Trajectory Subsampling}

This method matched standard PPO performance and in several cases exceeded it in stability.

\paragraph{Key insight.} Methods 1 and 2 both intervened at the data collection stage, before advantage estimation runs, which damaged reward signal integrity. Method 3 intervenes at the optimisation stage, after advantage estimation, so the ground truth signal is fully preserved while the gradient updates are still decorrelated.

\paragraph{Procedure.} The full trajectory buffer is collected normally with no skipping. Advantages are computed over the complete, unmodified transition sequence:
\begin{equation}
\hat{A}_t = \sum_{l=0}^{\infty} (\gamma \lambda)^l \delta_{t+l}, \quad \delta_t = r_t + \gamma V(s_{t+1}) - V(s_t)
\end{equation}
After this, a randomly chosen $p\%$ of the $N$ transitions is sampled without replacement for the gradient update. The remaining $(1-p)\%$ are excluded only from the optimisation step. Their reward contributions are already captured in the advantage estimates.

\paragraph{Theoretical justification.} The idea borrows from Dropout \cite{srivastava2014dropout}, not the mechanism but the logic. Dropout randomly kills neurons during the forward pass to stop the network from co-adapting to redundant features. Here the same principle is applied one level higher: instead of dropping neurons, we drop transitions. Randomly removing correlated transitions from the gradient update stops the optimizer from repeatedly reinforcing the same near-collinear gradient directions that on-policy trajectories naturally produce. The intervention point is different, Dropout sits inside the network while subsampling sits on the batch, but the principle is the same: inject controlled randomness to break correlated signal pathways before they corrupt optimisation.

Three things follow naturally from this design:
\begin{enumerate}
    \item \textbf{Decorrelation:} Randomly selecting which transitions enter the gradient update disrupts the sequential structure of the trajectory directly. No reward information is lost, just the redundant repetition of similar gradients.
    \item \textbf{Memory efficiency:} Pushing only $p\%$ of transitions to the GPU per update lowers memory pressure, which in practice means cleaner, less noisy gradient computation.
    \item \textbf{Implicit regularization:} Because the optimizer never sees the full correlated batch the same way twice, it cannot overfit to the local redundancy of any single trajectory, which nudges it toward policies that generalise better across state space.
\end{enumerate}

\begin{algorithm}[H]
\caption{PPO with Random $p\%$ Transition Subsampling}
\label{alg:ppo-subsample}
\begin{algorithmic}[1]
    \Require Policy $\pi_\theta$, critic $V_\phi$, rollout length $T$, subsampling fraction $p \in (0,1]$, clip $\varepsilon$, discount $\gamma$, GAE parameter $\lambda$
    \For{each iteration $k = 1, 2, \dots$}
        \State Execute $\pi_{\theta_k}$ for $T$ timesteps; collect buffer $\mathcal{D} = \{(s_t, a_t, r_t, s_{t+1}, d_t)\}_{t=0}^{T-1}$
        \State Compute TD residuals $\delta_t = r_t + \gamma(1 - d_t)V_\phi(s_{t+1}) - V_\phi(s_t)$ for all $t$
        \State Compute GAE advantages $\hat{A}_t = \sum_{l=0}^{T-1-t}(\gamma\lambda)^l \delta_{t+l}$ and returns $\hat{R}_t = \hat{A}_t + V_\phi(s_t)$
        \State Normalise: $\hat{A}_t \leftarrow \bigl(\hat{A}_t - \bar{A}\bigr) \big/ \bigl(\mathrm{std}(\hat{A}) + \epsilon\bigr)$
        \State Draw subsampled index set $\mathcal{I} \subset \{0,\dots,T-1\}$, $|\mathcal{I}| = \lfloor p \cdot T \rfloor$, sampled uniformly without replacement
        \For{each optimisation epoch}
            \State Partition $\mathcal{I}$ into minibatches $\{\mathcal{M}_j\}$
            \For{each minibatch $\mathcal{M}_j$}
                \State Compute probability ratio $\rho_t = \pi_\theta(a_t \mid s_t) \,/\, \pi_{\theta_k}(a_t \mid s_t)$
                \State $\mathcal{L}^{\text{CLIP}}(\theta) = \mathbb{E}_{t \in \mathcal{M}_j}\!\left[\min\!\left(\rho_t \hat{A}_t,\ \mathrm{clip}(\rho_t, 1\!-\!\varepsilon, 1\!+\!\varepsilon)\,\hat{A}_t\right)\right]$
                \State $\mathcal{L}^{\text{VF}}(\phi) = \mathbb{E}_{t \in \mathcal{M}_j}\!\left[\bigl(V_\phi(s_t) - \hat{R}_t\bigr)^2\right]$
                \State Update $\theta \leftarrow \theta + \alpha_\pi \nabla_\theta \mathcal{L}^{\text{CLIP}}(\theta)$ and $\phi \leftarrow \phi - \alpha_V \nabla_\phi \mathcal{L}^{\text{VF}}(\phi)$
            \EndFor
        \EndFor
    \EndFor
\end{algorithmic}
\end{algorithm}

\paragraph{On early training and convergence behaviour.}
In the early stages of training, the dropped and undropped agents do not look meaningfully different in terms of the actions they take, and that is expected. Early on the policy is essentially random regardless. Whether you drop 25\% of transitions or keep all of them, the agent is still exploring near-uniformly and the loss is high either way. The real difference shows up later. As training progresses and backpropagation starts shaping the network's internal representations, the subsampled agent begins generalising across states more smoothly, because its gradient updates were never locked into the redundant sequential structure of any single rollout. Given enough training steps both agents reach comparable final reward, but the subsampled agent tends to get there more stably. The reward is the last metric to reflect an improvement in training dynamics. The real signal is in KL divergence and entropy, and that is exactly where Method 3 shows its edge.

\section{Experimental Setup}

\subsection{Environments}
Experiments were run on five benchmark environments of increasing complexity:
\begin{itemize}
    \item \textbf{CartPole-v1} — A pole balanced on a cart moving along a frictionless track. The agent pushes left or right to stop it falling. Rewards come every timestep the pole stays up, the state is just 4 numbers, and the whole thing runs fast. It is the simplest possible sanity check — if a method cannot work here, nothing else matters.
    \item \textbf{Acrobot-v1} — A two-link pendulum where the agent can only apply torque at the middle joint, and has to swing the free end up past a target height. Unlike CartPole, there is no reward for progress — just a penalty every timestep until the goal is reached. That sparsity makes credit assignment noticeably harder and exposes weaknesses that CartPole would never catch.
    \item \textbf{LunarLander-v2} — A lander that needs to touch down safely between two flags using main and side thrusters. The reward signal tracks position, velocity, tilt, leg contact, and fuel use all at once, which makes it a genuinely difficult shaped-reward problem. This is the environment where long-horizon credit assignment actually matters, and where Methods 1 and 2 are expected to struggle.
    \item \textbf{HalfCheetah-v5} — A planar two-legged robot that must learn to run forward as fast as possible. The state space is high-dimensional and the reward is dense but shaped around velocity, making it a strong test of whether the subsampling method holds up under continuous control with a rich observation space.
    \item \textbf{Hopper-v5} — A single-legged robot that must learn to hop forward without falling over. Despite having fewer degrees of freedom than HalfCheetah, Hopper is notoriously sensitive to instability — small policy errors compound quickly and cause the agent to fall. It tests whether Method 3 can maintain stable training under the kind of fragile dynamics that tend to amplify any noise in the gradient.
\end{itemize}

\subsection{Algorithm}
All experiments used PPO with 1400 rollout steps per update. The baseline is standard PPO trained on the full trajectory buffer with no subsampling.

\subsection{Hyperparameters}

Tables~\ref{tab:hyperparams-discrete} and~\ref{tab:hyperparams-continuous} report the full configuration for every run. \textbf{Pure PPO} refers to $p = 100\%$ (no subsampling); \textbf{Skip-K} and \textbf{Rand-Skip} refer to Method~1 and Method~2 respectively.

\begin{table}[h]
\centering
\caption{Hyperparameters for CartPole-v1, Acrobot-v1, and LunarLander-v2.}
\label{tab:hyperparams-discrete}
\renewcommand{\arraystretch}{1.3}
\resizebox{\textwidth}{!}{%
\begin{tabular}{l cccc ccc ccc}
\toprule

& \multicolumn{4}{c}{\textbf{CartPole-v1}}
& \multicolumn{3}{c}{\textbf{Acrobot-v1}}
& \multicolumn{3}{c}{\textbf{LunarLander-v2}} \\

\cmidrule(lr){2-5} \cmidrule(lr){6-8} \cmidrule(lr){9-11}

\textbf{Hyperparameter}
& Pure PPO & $p = 75\%$ & Skip-K & Rand-Skip
& Pure PPO & $p = 80\%$ & $p = 65\%$
& Pure PPO & $p = 75\%$ & $p = 65\%$ \\

\midrule

Max Training Steps
& 500K & 500K & 500K & 500K
& 900K & 900K & 900K
& 1M & 1M & 1M \\

Rollout Steps
& 1400 & 1400 & 1400 & 1400
& 1400 & 1400 & 1400
& 1400 & 1400 & 1400 \\

PPO Clip ($\varepsilon$)
& 0.20 & 0.20 & 0.20 & 0.20
& 0.20 & 0.20 & 0.20
& 0.18 & 0.18 & 0.18 \\

Entropy Coeff.\ ($\beta_0$)
& 0.01 & 0.01 & 0.01 & 0.01
& 0.09 & 0.09 & 0.09
& 0.05 & 0.05 & 0.05 \\

Optimizer
& AdamW & AdamW & AdamW & AdamW
& AdamW & AdamW & AdamW
& AdamW & AdamW & AdamW \\

Actor LR
& 3e-4 & 3e-4 & 3e-4 & 3e-4
& 3e-4 & 3e-4 & 3e-4
& 3e-4 & 3e-4 & 3e-4 \\

Critic LR
& 5e-4 & 5e-4 & 5e-4 & 5e-4
& 5e-4 & 5e-4 & 5e-4
& 5e-4 & 5e-4 & 5e-4 \\

Epochs per Update
& 1 & 1 & 1 & 1
& 1 & 1 & 1
& 1 & 1 & 1 \\

Actor Params
& 17,026 & 17,026 & 17,026 & 17,026
& 17,219 & 17,219 & 17,219
& 17,412 & 17,412 & 17,412 \\

Critic Params
& 16,961 & 16,961 & 16,961 & 16,961
& 16,089 & 16,089 & 16,089
& 17,217 & 17,217 & 17,217 \\

$\gamma$
& 0.99 & 0.99 & 0.99 & 0.99
& 0.99 & 0.99 & 0.99
& 0.99 & 0.99 & 0.99 \\

GAE $\lambda$
& 0.98 & 0.98 & 0.98 & 0.98
& 0.98 & 0.98 & 0.98
& 0.98 & 0.98 & 0.98 \\

\bottomrule
\end{tabular}%
}
\end{table}

\begin{table}[h]
\centering
\small
\caption{Hyperparameters for HalfCheetah-v5 and Hopper-v5 (shared configuration). Epochs per update and hidden dimension $d_\text{model}$ are reduced at lower $p$ to match the smaller effective batch.}
\label{tab:hyperparams-continuous}
\renewcommand{\arraystretch}{1.2}
\begin{tabular}{lcccc}
\toprule
\textbf{Hyperparameter}
& \textbf{Pure PPO}
& $p = 85\%$
& $p = 75\%$
& $p = 65\%$ \\
\midrule
Max Training Steps          & 1.2M   & 1.2M   & 1.2M   & 1.2M   \\
Rollout Steps               & 2048   & 2048   & 2048   & 2048   \\
PPO Clip ($\varepsilon$)    & 0.18   & 0.18   & 0.18   & 0.15   \\
Optimizer                   & Adam   & Adam   & Adam   & Adam   \\
Actor LR                    & 9e-5   & 9e-5   & 9e-5   & 9e-5   \\
Critic LR                   & 3e-4   & 3e-4   & 3e-4   & 3e-4   \\
Epochs per Update           & 10     & 10     & 8      & 6      \\
Batch Size                  & 256    & 256    & 256    & 256    \\
$d_\text{model}$            & 128    & 128    & 96     & 96     \\
$\gamma$                    & 0.995  & 0.995  & 0.995  & 0.995  \\
GAE $\lambda$               & 0.96   & 0.96   & 0.96   & 0.96   \\
\bottomrule
\end{tabular}
\end{table}

\noindent
The following apply to HalfCheetah-v5 and Hopper-v5 across all runs: actor and critic gradient norms are clipped at 0.5 and 0.8 respectively; the entropy coefficient $\beta$ decays linearly from $\beta_0 = 1 \times 10^{-4}$ to zero over the course of training. A rollout length of 2048 was chosen because both environments have an average episode length of approximately 1000 steps — training on a single environment with a shorter buffer would produce gradient estimates too noisy to learn stable locomotion policies, so two effective environments worth of experience is collected per update. All forward passes for these two environments were executed under mixed-precision (FP16) using \texttt{torch.autocast}, with gradient scaling applied separately to the actor and critic to maintain numerical stability.

\subsection{Compute}

All experiments were run on a single NVIDIA Tesla T4 GPU (16\,GB VRAM) provided through Kaggle's free compute tier. No paid cloud resources were used. Tables~\ref{tab:time-discrete} and~\ref{tab:time-continuous} report wall-clock training times averaged over 5 independent seeds, as recorded by the Weights \& Biases run dashboard.

\begin{table}[h]
\centering
\caption{Average runtime per run for CartPole-v1, Acrobot-v1, and LunarLander-v2 (T4 GPU).}
\label{tab:time-discrete}
\renewcommand{\arraystretch}{1.3}
\resizebox{\textwidth}{!}{%
\begin{tabular}{l cccc ccc ccc}
\toprule

& \multicolumn{4}{c}{\textbf{CartPole-v1}}
& \multicolumn{3}{c}{\textbf{Acrobot-v1}}
& \multicolumn{3}{c}{\textbf{LunarLander-v2}} \\

\cmidrule(lr){2-5} \cmidrule(lr){6-8} \cmidrule(lr){9-11}

& Pure PPO & $p = 75\%$ & Skip-K & Rand-Skip
& Pure PPO & $p = 80\%$ & $p = 65\%$
& Pure PPO & $p = 75\%$ & $p = 65\%$ \\

\midrule

Avg.\ Runtime (mins)
& $\approx 22.8$ & $\approx 20.3$ & $\approx 21.6$ & $\approx 20.1$
& $\approx 28.2$ & $\approx 27.8$ & $\approx 25.4$
& $\approx 43.4$ & $\approx 41.7$ & $\approx 40.03$ \\

\bottomrule
\end{tabular}%
}
\end{table}

\begin{table}[h]
\centering
\caption{Average runtime per run for HalfCheetah-v5 and Hopper-v5 (T4 GPU).}
\label{tab:time-continuous}
\renewcommand{\arraystretch}{1.3}
\begin{tabular}{lcccc}
\toprule
& \textbf{Pure PPO} ($p = 100\%$) & $p = 85\%$ & $p = 75\%$ & $p = 65\%$ \\
\midrule
HalfCheetah-v5 (mins) & $\approx 38.7$ & $\approx 38.5$ & $\approx 37.5$ & $\approx 32.0$ \\
Hopper-v5 (mins)      & $\approx 43.3$ & $\approx 42.6$ & $\approx 40.7$ & $\approx 39.5$ \\
\bottomrule
\end{tabular}
\end{table}

\subsection{Evaluation Protocol}
Each run was trained across 5 independent seeds. At each evaluation checkpoint, the agent was evaluated for 5 episodes every \(K\) training steps, and the mean score across those 5 episodes was recorded. The reported reward is the mean across all 5 seeds, which smooths environment stochasticity and provides a reliable measure of policy performance.

\subsection{Evaluation Metrics}
The following metrics were tracked throughout training: KL divergence, policy entropy, explained variance, value bias, critic loss, and evaluation reward. Training stability was assessed by comparing the variance of these metrics across updates between Method~3 and the vanilla PPO baseline.

\section{Results}

Method 3 matched vanilla PPO on all five environments. Even after dropping $(1-p)\%$ of transitions per rollout, KL divergence, policy loss, value loss, and evaluation reward all stayed nearly identical to the baseline — and in several cases were measurably more stable, with lower variance across updates. The takeaway is that a large fraction of transitions in a typical on-policy rollout are simply redundant. Cutting them out randomly does not hurt anything; if anything, it regularises the gradient.

Methods 1 and 2 only held together on CartPole-v1. Acrobot-v1 exposed early instability in both, and LunarLander-v2 broke them outright. This is consistent with the central argument: the intervention point is what matters, and both methods got it wrong by touching the data before advantage estimation had a chance to preserve the reward signal.

On HalfCheetah-v5 and Hopper-v5, Method~3 tracked vanilla PPO closely on reward, KL divergence, and entropy across all tested $p$ values. Hopper in particular showed reduced metric variance under subsampling, which is notable given how sensitive that environment is to policy instability.

\paragraph{On the choice of $p$.}
At $p = 75\%$ all tracked metrics remained healthy across every environment — reward, entropy, and KL all matched vanilla PPO throughout training. Below $75\%$ the reward curve still looks fine, but entropy starts drifting and KL gets noisier: the optimizer is quietly losing the signal diversity it needs for stable exploration. The reward is the last metric to break. $p = 75\%$ is where all metrics still agree, which is why it is treated as the recommended threshold.

\section{Discussion}

What these experiments show, taken together, is that temporal correlation in on-policy trajectories should be addressed after advantage estimation, not before or during data collection. Intervening before advantage estimation destroys the reward signal the agent depends on for credit assignment. Intervening after leaves it completely intact, while the random subsampling of transitions for the weight update introduces enough stochasticity to break the correlated gradient structure.

A reasonable objection here is that standard PPO already shuffles rollout data into minibatches before each gradient update, so doesn't that handle the correlation? It does not. Shuffling changes the order transitions arrive in but does not remove any of them. The core problem is that the states within a single on-policy rollout are highly similar to one another — they follow causally from the same policy acting in the same environment over a short window of time. Feeding all of them into the optimizer, regardless of order, still pushes the gradient repeatedly in nearly the same direction. Subsampling to $p\%$ actively discards some of that redundant overlap, which is precisely why it reduces gradient collinearity in a way that shuffling cannot. The key result here is not just parity with vanilla PPO but the fact that a strict subset of the rollout transitions is sufficient to recover identical final performance — which is direct evidence that the full correlated batch was contributing less unique gradient signal than its size implied.

Method 3 also does not touch anything it should not. Methods 1 and 2 both accumulated rewards across skipped transitions, which quietly broke the Markov assumption — the stored $s_t$ ended up carrying information from future states it never actually observed. Method 3 has no such side effect. The environment, the rollout, advantage estimation, and the clipping objective are all identical to vanilla PPO. The only change is a single random sampling step between advantage estimation and the gradient update.

\paragraph{On network size, learning rates, and gradient clipping.}
For the continuous control environments, keeping the network deliberately small turned out to matter more than it might seem. A larger hidden dimension with more epochs per update causes the critic to overfit quickly to the current batch. The MSE loss drops fast and then starts oscillating as the value estimates chase a moving target. A smaller network generalises more smoothly across the state distribution and stays stable for longer. The learning rate asymmetry between actor and critic follows the same logic. The critic is trained via MSE and converges much faster, so a higher critic learning rate is fine and actually helps it track the returns more accurately. The actor, by contrast, needs to change slowly because aggressive policy updates in locomotion tasks compound into instability very quickly. Keeping the actor learning rate lower ensures that each policy step is a small, grounded improvement rather than an overcorrection. The critic gradient norm being clipped at 0.8 compared to the actor at 0.5 reflects the same asymmetry. The critic gradient can grow large during early training when value estimates are far off, and clipping it at a slightly higher threshold lets it learn faster without exploding, while the actor is kept on a tighter leash throughout.

\section{Conclusion}

On-policy rollouts carry more redundancy than is commonly assumed. The experiments here show that randomly dropping $25\%$ of transitions before the gradient update — at the right stage so the reward signal stays intact — is enough to break the correlated gradient structure and stabilise training, without touching anything else in the PPO pipeline. The method matches standard PPO on evaluation reward across five environments while consistently reducing variance in the training metrics that matter most.

The result was stronger than expected. Despite dropping $(1-p)\%$ of transitions per update, evaluation returns after sufficient training are nearly identical to the undropped baseline and often slightly better. This is not a coincidence — it is direct empirical evidence that the states within a rollout are so highly correlated that a large fraction of them contribute little unique information to the gradient. The network was already learning from redundant signal. Removing some of it does not hurt; it cleans things up.

On the choice of $p$: at $75\%$ all tracked metrics still agree — reward, entropy, and KL divergence all healthy. Below that the reward holds but entropy drifts and KL gets noisier, and the optimizer is quietly losing grip on stable exploration before the reward has reflected it. Dropping exactly $25\%$ is the practical sweet spot.

\bibliographystyle{plain}

\newpage

\section*{Appendix: Experimental Results and Graphs}

All figures compare Vanilla PPO, Method 1 (Fixed K-Step), Method 2 (Random Adaptive K-Step), and Method 3 (Random $p\%$ Subsampling) across 1400 rollout steps per update. HalfCheetah-v5 and Hopper-v5 compare Method 3 against the vanilla PPO baseline.

\subsection*{CartPole-v1 (Fig.\ 1--6, $p = 75\%$)}

\begin{figure}[h]
    \centering
    \begin{minipage}{0.48\linewidth}
        \centering
        \includegraphics[width=\linewidth]{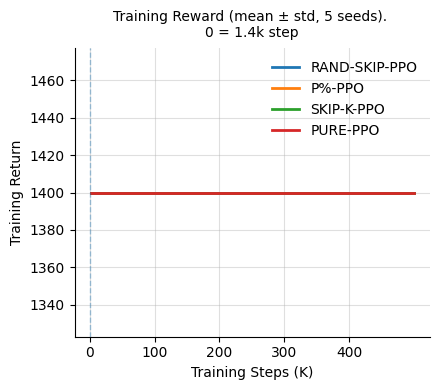}
        \caption{CartPole-v1: Training reward.}
    \end{minipage}
    \hfill
    \begin{minipage}{0.48\linewidth}
        \centering
        \includegraphics[width=\linewidth]{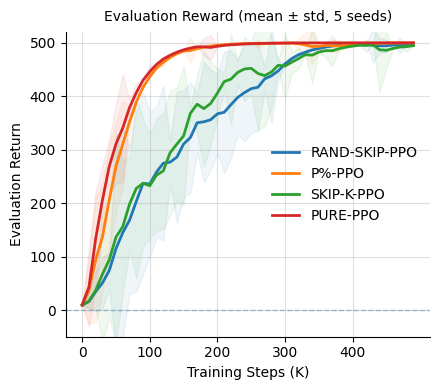}
        \caption{CartPole-v1: Evaluation reward.}
    \end{minipage}
\end{figure}

\begin{figure}[h]
    \centering
    \begin{minipage}{0.48\linewidth}
        \centering
        \includegraphics[width=\linewidth]{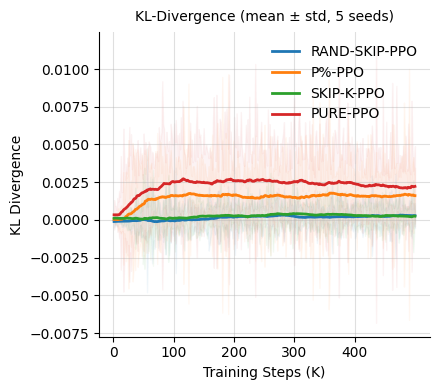}
        \caption{CartPole-v1: KL divergence.}
    \end{minipage}
    \hfill
    \begin{minipage}{0.48\linewidth}
        \centering
        \includegraphics[width=\linewidth]{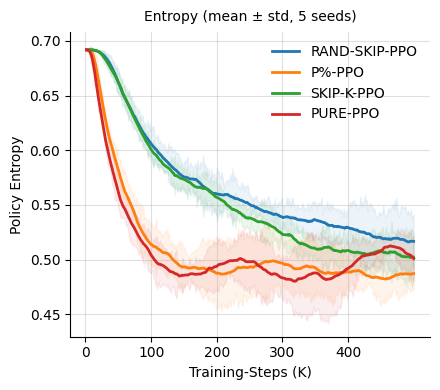}
        \caption{CartPole-v1: Policy entropy.}
    \end{minipage}
\end{figure}

\begin{figure}[h]
    \centering
    \begin{minipage}{0.48\linewidth}
        \centering
        \includegraphics[width=\linewidth]{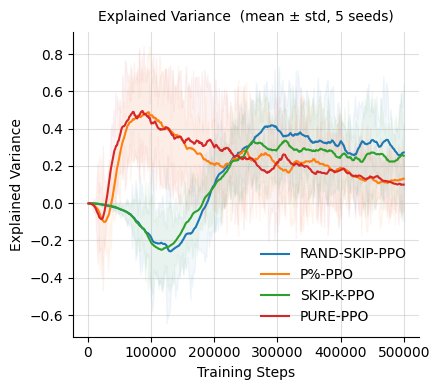}
        \caption{CartPole-v1: Explained variance.}
    \end{minipage}
    \hfill
    \begin{minipage}{0.48\linewidth}
        \centering
        \includegraphics[width=\linewidth]{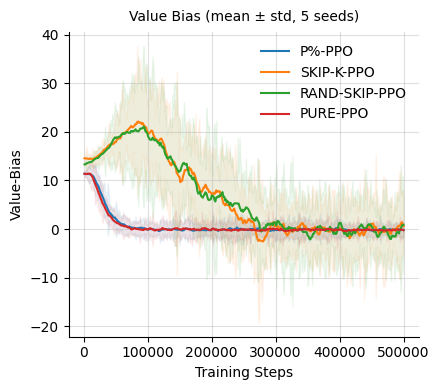}
        \caption{CartPole-v1: Value bias.}
    \end{minipage}
\end{figure}

\subsection*{Acrobot-v1 (Fig.\ 7--12)}

\begin{figure}[h]
    \centering
    \begin{minipage}{0.48\linewidth}
        \centering
        \includegraphics[width=\linewidth]{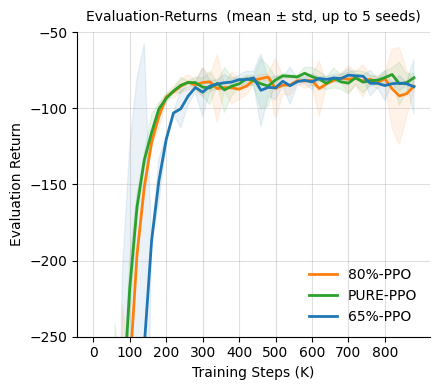}
        \caption{Acrobot-v1: Evaluation reward.}
    \end{minipage}
    \hfill
    \begin{minipage}{0.48\linewidth}
        \centering
        \includegraphics[width=\linewidth]{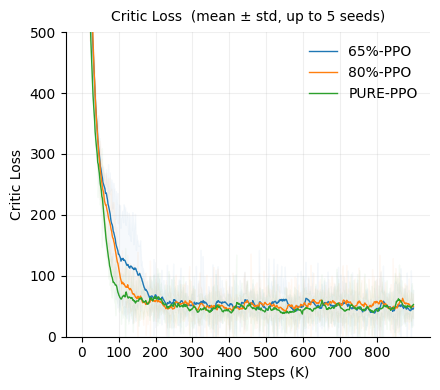}
        \caption{Acrobot-v1: Critic loss.}
    \end{minipage}
\end{figure}

\begin{figure}[h]
    \centering
    \begin{minipage}{0.48\linewidth}
        \centering
        \includegraphics[width=\linewidth]{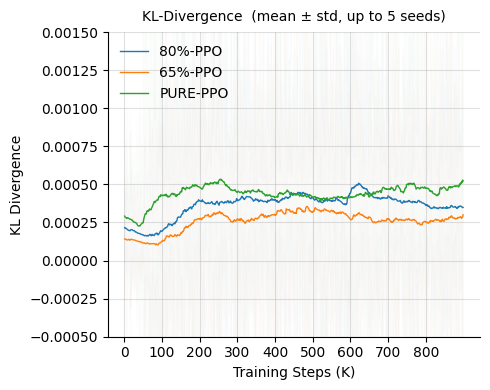}
        \caption{Acrobot-v1: KL divergence.}
    \end{minipage}
    \hfill
    \begin{minipage}{0.48\linewidth}
        \centering
        \includegraphics[width=\linewidth]{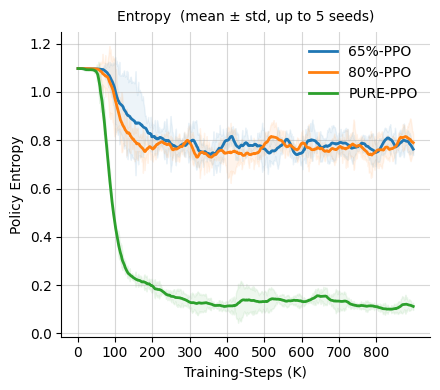}
        \caption{Acrobot-v1: Policy entropy.}
    \end{minipage}
\end{figure}

\begin{figure}[h]
    \centering
    \begin{minipage}{0.48\linewidth}
        \centering
        \includegraphics[width=\linewidth]{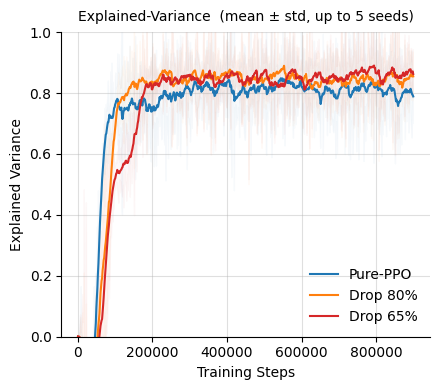}
        \caption{Acrobot-v1: Explained variance.}
    \end{minipage}
    \hfill
    \begin{minipage}{0.48\linewidth}
        \centering
        \includegraphics[width=\linewidth]{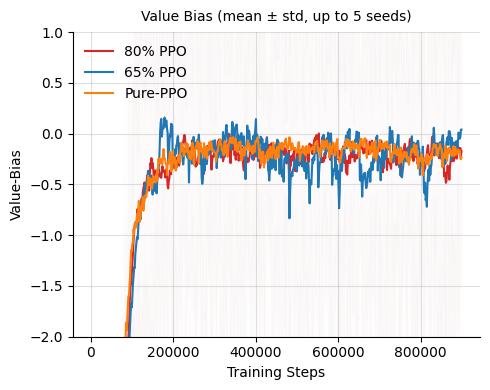}
        \caption{Acrobot-v1: Value bias.}
    \end{minipage}
\end{figure}

\subsection*{LunarLander-v2 (Fig.\ 13--18)}

\begin{figure}[h]
    \centering
    \begin{minipage}{0.48\linewidth}
        \centering
        \includegraphics[width=\linewidth]{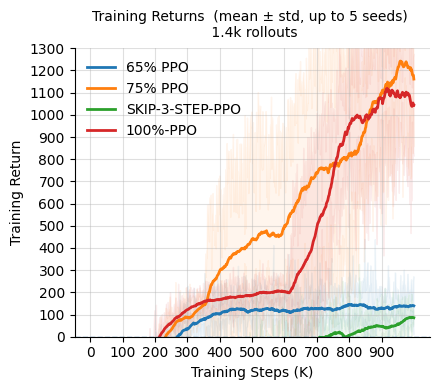}
        \caption{LunarLander-v2: Training reward.}
    \end{minipage}
    \hfill
    \begin{minipage}{0.48\linewidth}
        \centering
        \includegraphics[width=\linewidth]{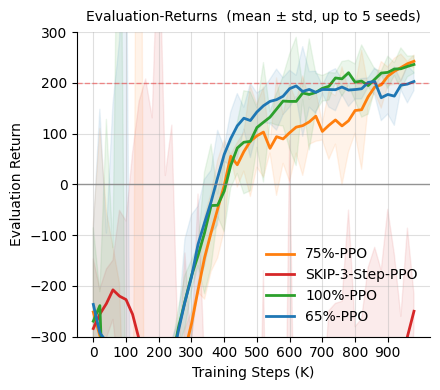}
        \caption{LunarLander-v2: Evaluation reward.}
    \end{minipage}
\end{figure}

\begin{figure}[h]
    \centering
    \begin{minipage}{0.48\linewidth}
        \centering
        \includegraphics[width=\linewidth]{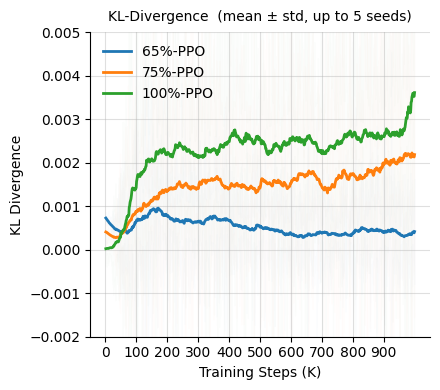}
        \caption{LunarLander-v2: KL divergence.}
    \end{minipage}
    \hfill
    \begin{minipage}{0.48\linewidth}
        \centering
        \includegraphics[width=\linewidth]{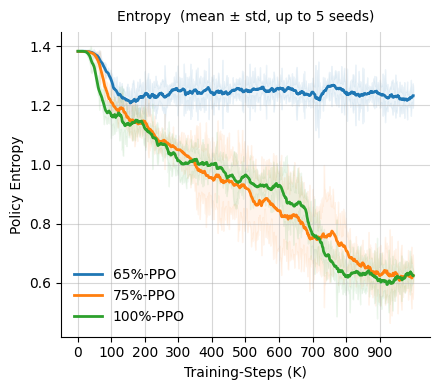}
        \caption{LunarLander-v2: Policy entropy.}
    \end{minipage}
\end{figure}

\begin{figure}[h]
    \centering
    \begin{minipage}{0.48\linewidth}
        \centering
        \includegraphics[width=\linewidth]{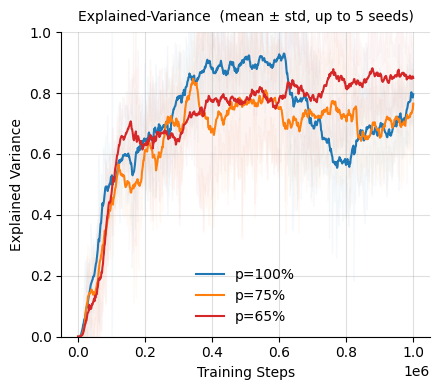}
        \caption{LunarLander-v2: Explained variance.}
    \end{minipage}
    \hfill
    \begin{minipage}{0.48\linewidth}
        \centering
        \includegraphics[width=\linewidth]{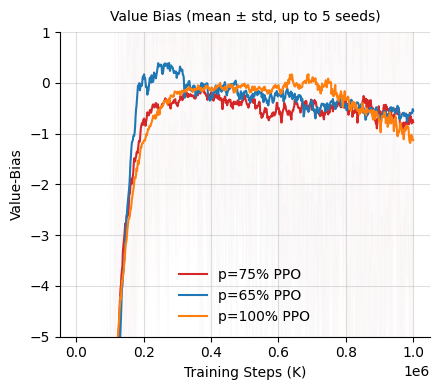}
        \caption{LunarLander-v2: Value bias.}
    \end{minipage}
\end{figure}

\subsection*{HalfCheetah-v5 (Fig.\ 19--24)}

\begin{figure}[h]
    \centering
    \begin{minipage}{0.48\linewidth}
        \centering
        \includegraphics[width=\linewidth]{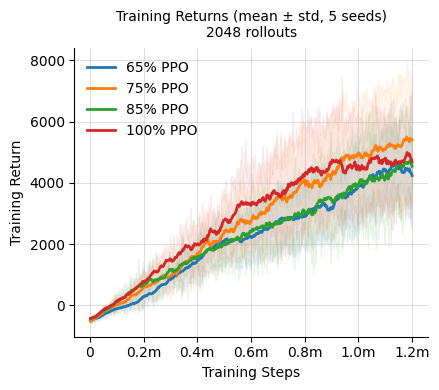}
        \caption{HalfCheetah-v5: Training reward.}
    \end{minipage}
    \hfill
    \begin{minipage}{0.48\linewidth}
        \centering
        \includegraphics[width=\linewidth]{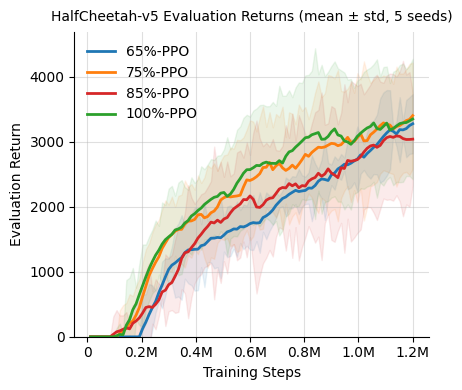}
        \caption{HalfCheetah-v5: Evaluation reward.}
    \end{minipage}
\end{figure}

\begin{figure}[h]
    \centering
    \begin{minipage}{0.48\linewidth}
        \centering
        \includegraphics[width=\linewidth]{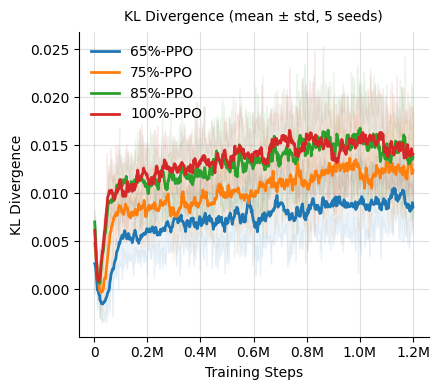}
        \caption{HalfCheetah-v5: KL divergence.}
    \end{minipage}
    \hfill
    \begin{minipage}{0.48\linewidth}
        \centering
        \includegraphics[width=\linewidth]{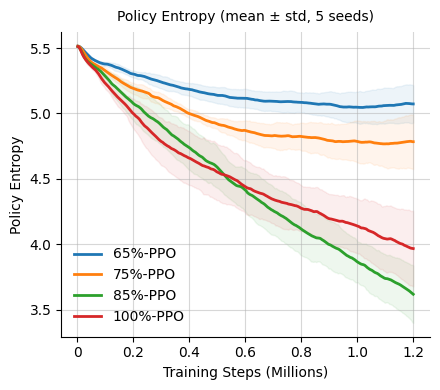}
        \caption{HalfCheetah-v5: Policy entropy.}
    \end{minipage}
\end{figure}

\begin{figure}[h]
    \centering
    \begin{minipage}{0.48\linewidth}
        \centering
        \includegraphics[width=\linewidth]{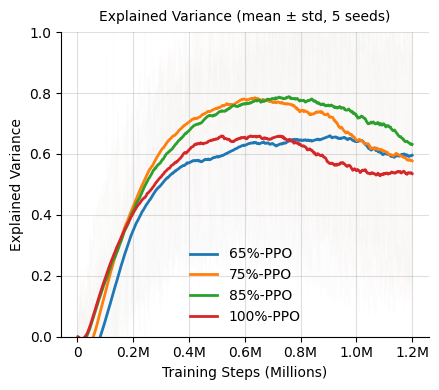}
        \caption{HalfCheetah-v5: Explained variance.}
    \end{minipage}
    \hfill
    \begin{minipage}{0.48\linewidth}
        \centering
        \includegraphics[width=\linewidth]{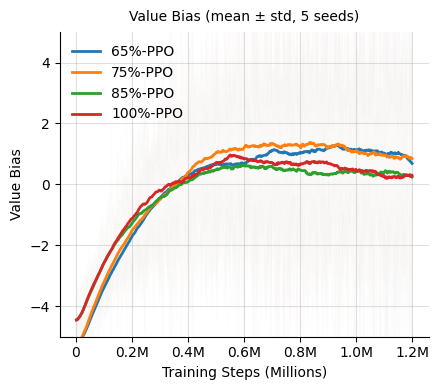}
        \caption{HalfCheetah-v5: Value bias.}
    \end{minipage}
\end{figure}

\begin{figure}[h]
    \centering
    \begin{minipage}{0.48\linewidth}
        \centering
        \includegraphics[width=\linewidth]{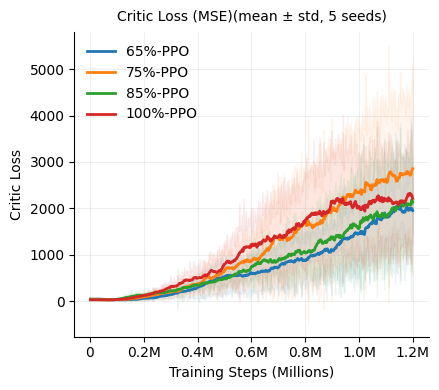}
        \caption{HalfCheetah-v5: Critic loss.}
    \end{minipage}
    \hfill
    \begin{minipage}{0.48\linewidth}
    \end{minipage}
\end{figure}

\subsection*{Hopper-v5 (Fig.\ 25--28)}

\begin{figure}[h]
    \centering
    \begin{minipage}{0.48\linewidth}
        \centering
        \includegraphics[width=\linewidth]{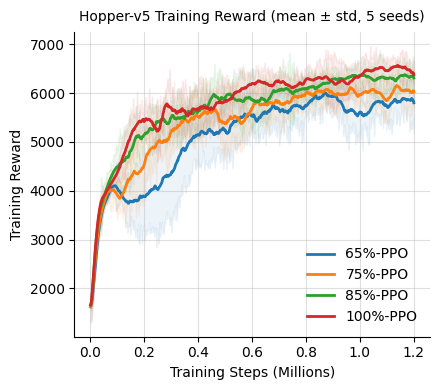}
        \caption{Hopper-v5: Training reward.}
    \end{minipage}
    \hfill
    \begin{minipage}{0.48\linewidth}
        \centering
        \includegraphics[width=\linewidth]{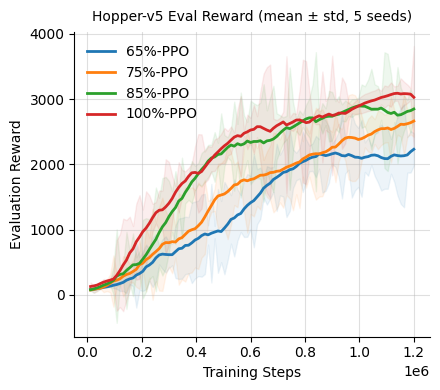}
        \caption{Hopper-v5: Evaluation reward.}
    \end{minipage}
\end{figure}

\begin{figure}[h]
    \centering
    \begin{minipage}{0.48\linewidth}
        \centering
        \includegraphics[width=\linewidth]{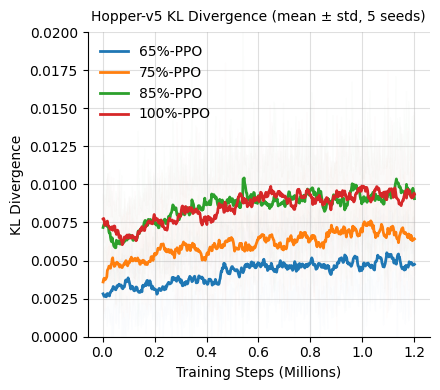}
        \caption{Hopper-v5: KL divergence.}
    \end{minipage}
    \hfill
    \begin{minipage}{0.48\linewidth}
        \centering
        \includegraphics[width=\linewidth]{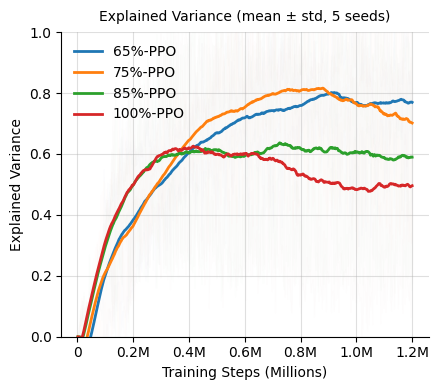}
        \caption{Hopper-v5: Explained variance.}
    \end{minipage}
\end{figure}

\end{document}